\documentclass[10pt,twocolumn,letterpaper]{article}
\usepackage{cvpr}
\usepackage{graphicx}
\usepackage{amsmath}
\usepackage{amssymb}
\usepackage{booktabs}
\usepackage{multirow}
\usepackage{pifont}
\newcommand{\xmark}{\ding{55}}%

\usepackage[table]{xcolor}
\definecolor{mygray}{gray}{.92}
\definecolor{mycyan}{cmyk}{.3,0,0,0}
\definecolor{LightCyan}{rgb}{0.88,1,1}

\makeatletter
\newcommand{\thickhline}{%
    \noalign {\ifnum 0=`}\fi \hrule height 1pt
    \futurelet \reserved@a \@xhline}

\usepackage[pagebackref,breaklinks,colorlinks]{hyperref}

\usepackage[capitalize]{cleveref}
\crefname{section}{Sec.}{Secs.}
\Crefname{section}{Section}{Sections}
\Crefname{table}{Table}{Tables}
\crefname{table}{Tab.}{Tabs.}

\def\cvprPaperID{4022} % *** Enter the CVPR Paper ID here
\def\confName{CVPR}
\def\confYear{2022}

\begin{document}

\title{Modulated Contrast for Versatile Image Synthesis}
% \title{Modulated Contrastive Learning for Versatile Image Translation}
% Coordinated 

\author{
Fangneng Zhan\textsuperscript{\rm 1},
Jiahui Zhang\textsuperscript{\rm 2},
Yingchen Yu\textsuperscript{\rm 2},
Rongliang Wu\textsuperscript{\rm 2}, 
Shijian Lu\thanks{Corresponding author, E-mail: shijian.lu@ntu.edu.sg} \ \textsuperscript{\rm 2}   \\
% \textsuperscript{\rm 1} 
\textsuperscript{\rm 1} S-Lab, Nanyang Technological University
\quad
\textsuperscript{\rm 2} Nanyang Technological University
% Academy, Alibaba Group
% Nanyang Technological University   
}

\maketitle

\begin{abstract}
Perceiving the similarity between images has been a long-standing and fundamental problem underlying various visual generation tasks. Predominant approaches measure the inter-image distance by computing pointwise absolute deviations, which tends to estimate the median of instance distributions and leads to blurs and artifacts in the generated images. This paper presents MoNCE, a versatile metric that introduces image contrast to learn a calibrated metric for the perception of multifaceted inter-image distances. Unlike vanilla contrast which indiscriminately pushes negative samples from the anchor regardless of their similarity,
we propose to re-weight the pushing force of negative samples adaptively according to their similarity to the anchor, which facilitates the contrastive learning from informative negative samples. Since multiple patch-level contrastive objectives are involved in image distance measurement, we introduce optimal transport in MoNCE to modulate the pushing force of negative samples collaboratively across multiple contrastive objectives. Extensive experiments over multiple image translation tasks show that the proposed MoNCE outperforms various prevailing metrics substantially. The code is available at \href{https://github.com/fnzhan/MoNCE}{MoNCE}.
\end{abstract}

\begin{figure}[t]
\centering
\includegraphics[width=1.0\linewidth]{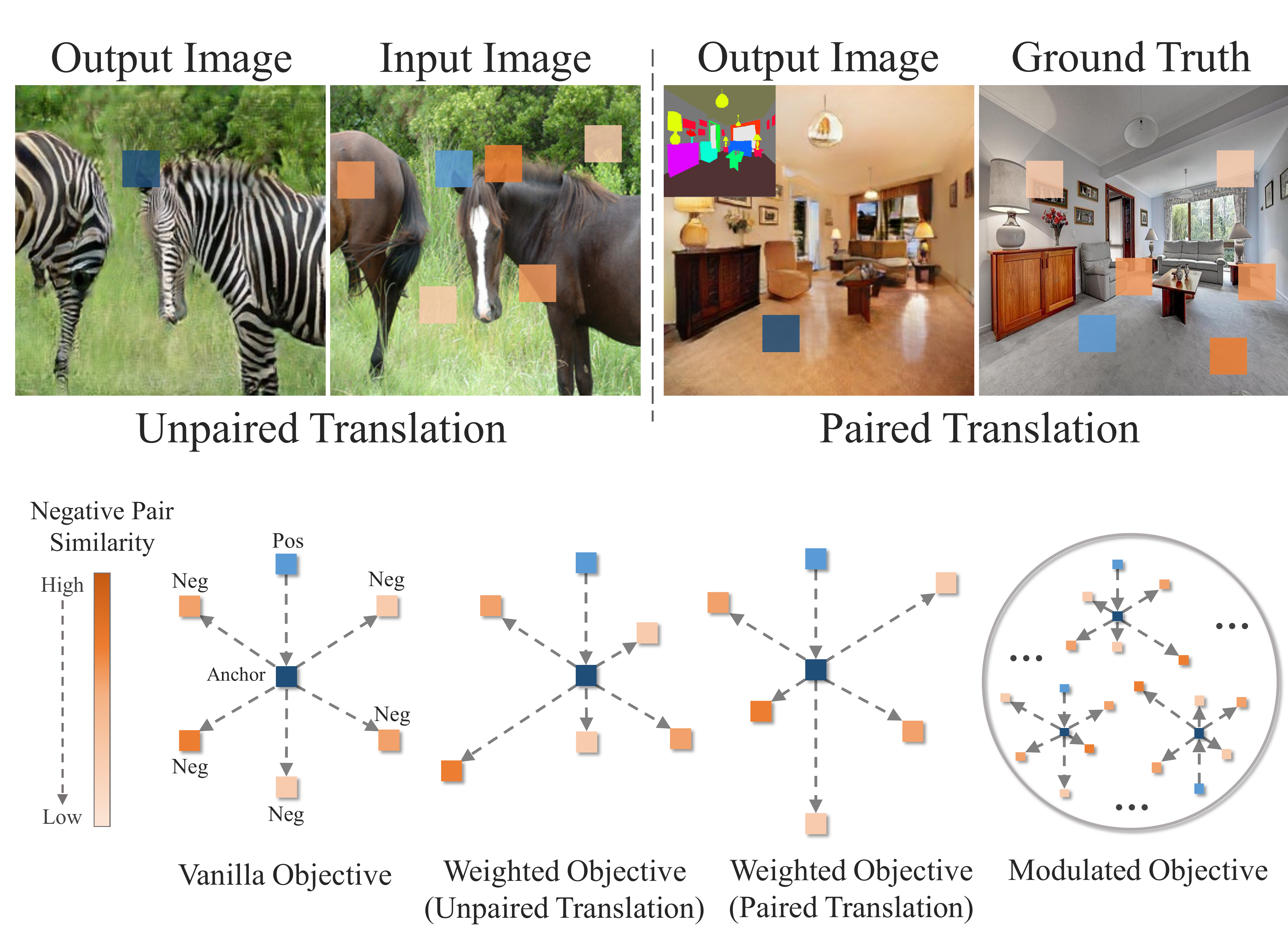}
\caption{
Comparison of different contrastive objectives:
For the contrastive objective of a single image patch,
vanilla contrastive objective repels all negative samples indiscriminately.
The introduced weighted contrastive objective adaptively adjusts the weights of negative pairs according to the pair similarity. With inverse weighting strategies for unpaired and paired translation tasks, the weighted objective can be applied to improve the generation performance substantially.
The modulated contrastive objective introduces optimal transport to modulate the learning objectives of all image patches as a whole.
}
\label{im_intro}
\end{figure}

\section{Introduction}

Multifarious image generation tasks \cite{zhu2017unpaired,park2019spade,park2020contrastive,shrivastava2017learning,zhan2019sfgan,zhan2021unite,zhan2021rabit,zhu2017multimodal} often entail multifaceted metrics to measure the inter-image similarity with regard to different properties such as image structures, image semantics and image perceptual realism, etc. 
Defining generic metrics to fulfil multiple objectives is challenging as different visual properties are usually entangled in pixels and the notion of visual similarity is often subjective. Image similarity measurement remains a very open research challenge in visual generation tasks.

To measure and minimize the content variation in unpaired image translation, Zhu \emph{et al.} \cite{zhu2017unpaired} design a cycle-consistency loss to ensure that input images can be recovered from the output images.
Different from unpaired image translation,
paired image translation entails certain metrics to measure the perceptual similarity between output images and ground truth.
Among various distance metrics \cite{wang2004image,wang2003multiscale}, 
perceptual loss \cite{johnson2016perceptual} emerges as a powerful metric in line with human perception by leveraging the internal activation of pre-trained networks.
However, above metrics are designed based on point-wise deviations, which undesirably minimize the average deviation to all possible instances. For example, a semantic map corresponds to numerous real images, minimizing the average deviation to all possible real images tends to produce blurred generation results.

Instead of minimizing the point-wise deviation, the prevailing contrastive learning \cite{chen2020simple, he2020momentum, wu2018unsupervised} aims to pull positive samples towards an anchor and push negative samples far away from it.
It has recently been adopted in image generation tasks for preserving image contents in unpaired image translation \cite{park2020contrastive}, perceiving image similarity in paired image translation \cite{andonian2021contrastive}, or serving as a contrastive regularization term in image dehazing \cite{wu2021contrastive}.
However, all these studies adopt the vanilla contrast that shares a critical constraint – negative samples are indiscriminately pushed away from the anchor regardless of their similarity to the anchor.
% as illustrated in Fig. \ref{im_intro}.

In this work, we formulate contrastive learning as a versatile metric for various image translation tasks as shown in Fig. \ref{im_intro}.
In unpaired image translation, contrastive learning allows to preserve image contents by maximizing the mutual information of corresponding patches \cite{park2020contrastive}.
In paired image translation, contrastive learning is employed to measure the perceptual similarity between images in line with human judgement, by leveraging pre-trained networks for feature extraction.
However, vanilla contrastive objective repels all negative samples indiscriminately, which is apparently sub-optimal as negative samples usually have different similarity with the anchor. Certain weighting strategy is desired to formulate more effective contrast by adaptively adjusting the pushing force of negative samples.

\begin{figure*}[t]
\centering
\includegraphics[width=1.0\linewidth]{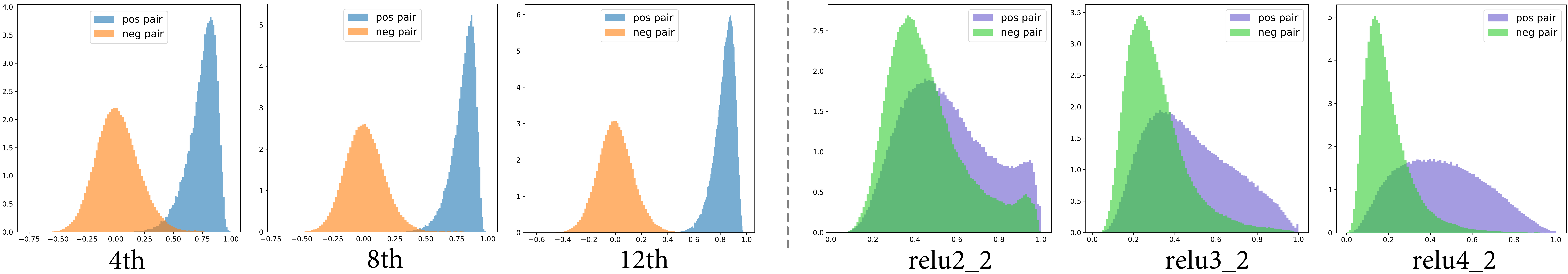}
\caption{
Histograms of positive \& negative pair similarity in unpaired and paired image translation.
The three plots on the left are the results of 4th, 8th, 12th layers of CUT model \cite{park2020contrastive} for unpaired image translation (Horse to Zebra).
The three plots on the right are the results of $relu2\_2,relu3\_2,relu4\_2$ layers of pre-trained VGG-19 in SPADE \cite{park2019spade} for paired image translation (ADE20K).
The very distinct similarity distribution leads to inverse weighting strategies for unpaired and paired image translation.
}
\label{im_distribution}
\end{figure*}

Aiming to boost the translation performance,
we comprehensively investigated different weighting strategies for negative samples and some non-trivial conclusions are drawn for the selection of weighting strategies in different scenarios.
Intuitively, hard negative samples (i.e., with high similarity to the anchor) should be assigned higher weights (referred as \textit{hard weighting}), complying with the rationale of hard negative sampling \cite{robinson2020contrastive}.
It is true for unpaired image translation where negative samples can be easily pushed apart as illustrated in the similarity distributions of negative \& positive pairs in Fig. \ref{im_distribution}.
However, for paired image translation, negative samples are hard to be pushed apart from the anchor (or positive pairs) as there is severe overlap for the similarity distribution of negative \& positive pairs as in Fig. \ref{im_distribution}.
% as illustrated in the similarity distribution of negative and positive pairs in Fig. \ref{im_distribution}.
In this scenario, we surprisingly find that the intuitive hard weighting strategy tends to impair the performance, and an inverse weighting strategy as shown in Fig. \ref{im_intro} allows to improve the performance.
In addition, as in PatchNCE loss \cite{park2020contrastive}, contrastive learning for measuring image similarity involves several sub-objectives as each image patch is associated with a contrastive objective. Re-weighting each sub-objective separately without overall coordination tends to be sub-optimal.
We propose a \textbf{Mo}dulated \textbf{N}oise \textbf{C}ontrastive \textbf{E}stimation (\textbf{MoNCE}) loss that employs optimal transport \cite{peyre2019computational} to modulate the re-weighting of all negative samples collaboratively across the multiple objectives.
With a cost matrix designed based on the similarity of negative pairs, optimal transport allows to retrieve an optimal transport plan which serves as the weights for negative samples to reach an overall optimal objective.

The contributions of this work can be summarized in three aspects. 
First, we formulate contrastive learning as a versatile metric in multifarious image translation tasks. 
Second, we extensively investigate the effect of negative pair weighting in contrastive learning and propose to adopt different weighting strategies according to the similarity distribution of negative pairs.
Third, we propose a modulated contrast that exploits optimal transport to modulate the re-weighting of all negative pairs collaboratively across multiple contrastive objectives.

\section{Related Work}

\paragraph{Image Generation Loss}

Image generation tasks entail various losses to achieve dedicated purposes in image synthesis \cite{mechrez2018contextual,mechrez2018maintaining,taigman2016unsupervised,wu2020cascade,wu2020leed,zhan2021emlight,zhan2021gmlight,yu2021diverse,yu2021wavefill,zhan2021multimodal}.
For instance, unpaired image translation is usually associated with certain losses to encourage correlation between the input and output images.
Specially, Zhu \emph{et al.} \cite{zhu2017unpaired} design a cycle-consistency loss to preserve the image content by ensuring the input image can be recovered from the translation result. 
However, cycle-consistency loss assumes the relationship between the two domains is a bijection which is often too restrictive for image translation tasks.
Therefore, several works \cite{benaim2017one,amodio2019travelgan,fu2019geometry} aim to explore one-way translation and bypass the bijection constraint of cycle-consistency.
At the other end, paired image translation entails certain metric to measure the perceptual similarity between images in line with human perception.
By leveraging the internal activation of pre-trained neural networks, perceptual loss \cite{johnson2016perceptual,dosovitskiy2016generating,gatys2016image,ulyanov2017improved} emerges as a powerful metric in image translation that coincides with human perception \cite{zhang2018lpips}.
However, all above metrics are designed based on point-wise absolute deviation, which tends to estimate the median of all possible instances.

With the emergence of contrastive learning, a popular line of research introduces contrastive learning in image generation \cite{deng2020disentangled,kang2020ContraGAN,yu2021dual,zhang2021cross,zhang2021blind}.
Specially, CUT \cite{park2020contrastive} proposes to maximize the mutual information between corresponding patches via noise contrastive estimation \cite{oord2018representation} for preserving the contents in unpaired image translation.
Andonian \emph{et al.} \cite{andonian2021contrastive} introduce contrastive learning to measure the inter-image similarity in paired image translation.
AECR-Net \cite{wu2021contrastive} introduces a contrastive regularization for image dehazing by pulling the restored image closer to the clear image and push it far away from the hazy image in the representation space.
NEGCUT \cite{wang2021instance} presents an instance-wise hard negative sample generation framework for contrastive learning in Unpaired image-to-image Translation.
However, all previous losses are designed based on the vanilla contrastive learning which indiscriminately repels all negative samples regardless of their similarity to the anchor.

\paragraph{Contrastive Learning}

The contrastive learning \cite{chen2020simple,he2020momentum,wu2018unsupervised} has recently become a prominent tool in unsupervised representation learning, leading to state-of-the-art results. 
The goal of contrastive learning (CL) is to learn a generic feature embedding by pulling positive points towards an anchor and push negative points far away from it.
However, the objective of conventional contrastive learning is misleading as negative samples will be pushed apart indiscriminately with the same weights regardless of their similarity to the anchor.
To alleviate the undesired repelling of similar pairs, a popular line of research explores to reweight the NCE loss by increasing the importance of positive pairs \cite{chuang2020debiased} or allocating different importance for negative pairs \cite{robinson2020contrastive}.
Besides, Chen \emph{et al.} \cite{chen2021large} propose large-margin contrastive learning (LMCL) to distinguish intra-cluster and inter-cluster pairs and only push away inter-cluster pair.
However, all described methods explore to re-weight a single contrastive objective for feature representation, which is infeasible for the cases accompanied with multiple contrastive objectives and cannot generalize to the area of image generation.

\section{Proposed Method}

In this section, we first formulate the contrastive learning as a versatile metric for unpaired and paired image translation tasks.
Then we establish the weighting strategies for unpaired and paired image translation according to the similarity distribution of positive and negative pairs.
Finally, we derive our designed modulated noise contrastive estimation (MoNCE) loss which enables to coordinate the re-weighting of negative pairs across multiple objectives.

\subsection{Versatile Metric for Image Translation}

Given images in two domains, image translation aims to translate images from the input domain to appear like images from the output domain.
The datasets for training translation model could be unpaired (i.e., unpaired image translation) and paired (i.e., paired image translation), and different loss terms are entailed for image translations with different dataset setting.
GAN loss is usually shared across unpaired and paired image translation to fight against artifacts in translated images, and other loss terms are usually designed specifically to fulfill various objectives, e.g., cycle loss \cite{zhu2017unpaired} for content preservation, perceptual loss \cite{johnson2016perceptual} for assessing human perceptual similarity.
However, most metrics are designed by computing the absolute mean error which tends to minimize the average deviation to all possible instances and leads to blurs in the generated images.

In this work, we formulate contrastive loss as a versatile metric in various translation tasks, just by properly selecting the positive and negative pairs.
For unpaired image translation, previously proposed PatchNCE \cite{park2020contrastive} has validated the effectiveness of contrastive learning for the preservation of content.
PatchNCE aims to maximize the mutual information between patches in the same spatial location from the generated image X and the ground truth Y as below:
\begin{equation}
\label{patchnce}
\mathcal{L}(X, Y) = -\sum_{i=1}^N \log
\frac{e^{x_i \cdot y_i / \tau}} 
{e^{x_i \cdot y_i / \tau} + \sum_{\substack{j=1 \\ j\neq i}}^N e^{ x_i \cdot y_j / \tau}},
\end{equation}
where $X = [x_1, x_2, \cdots, x_N]$ and $Y = [y_1, y_2, \cdots, y_N]$ are encoded image feature sets, $\tau$ is the temperature parameter, $N$ is the number of feature patches.
Normally, multi-layer features (1th, 4th, 8th, 12th and 16th layers of the encoder) are employed in PatchNCE, which is formulated as $\mathcal{L}^{m}(X,Y)= \sum_{l=1}^{L} \mathcal{L}(X_l, Y_l)$, where $X_l$ and $Y_l$ denote the corresponding feature sets in $l$ layer of the encoder.

For paired image translation, we aim to measure the perceptual similarity between translated images and the ground truth in line with human perception.
Consistent with the multi-layer setting in perceptual loss \cite{johnson2016perceptual}, we employ pre-trained VGG-19 network \cite{simonyan2014very} to extract the same layer features ($relu1\_2, relu2\_2, relu3\_2, relu4\_2, relu5\_2$) from translated images and ground truth to construct contrastive learning pairs.
By treating feature patches in same spatial location from the translated image and the ground truth as positive pairs and feature patches in different location as negative pairs, Eq. (\ref{patchnce}) can be utilized to maximize the mutual information between translated images and the ground truth.
According to the experimental results in Table \ref{tab_spade}, the PatchNCE with pre-trained VGG-19 for feature extraction rivals the well-known perceptual loss \cite{johnson2016perceptual} in terms of paired image translation in various evaluation metrics.

Considering the superior performance of PatchNCE for unpaired and paired image translation, contrastive learning can serve as a versatile metric in various image translation tasks. 
However, the vanilla objective of PatchNCE will repel all negative samples indiscriminately regardless of their similarity to the anchor, which tends to be sub-optimal as the inherent information of negative samples is not equal.

\begin{figure}[t]
\centering
\includegraphics[width=1.0\linewidth]{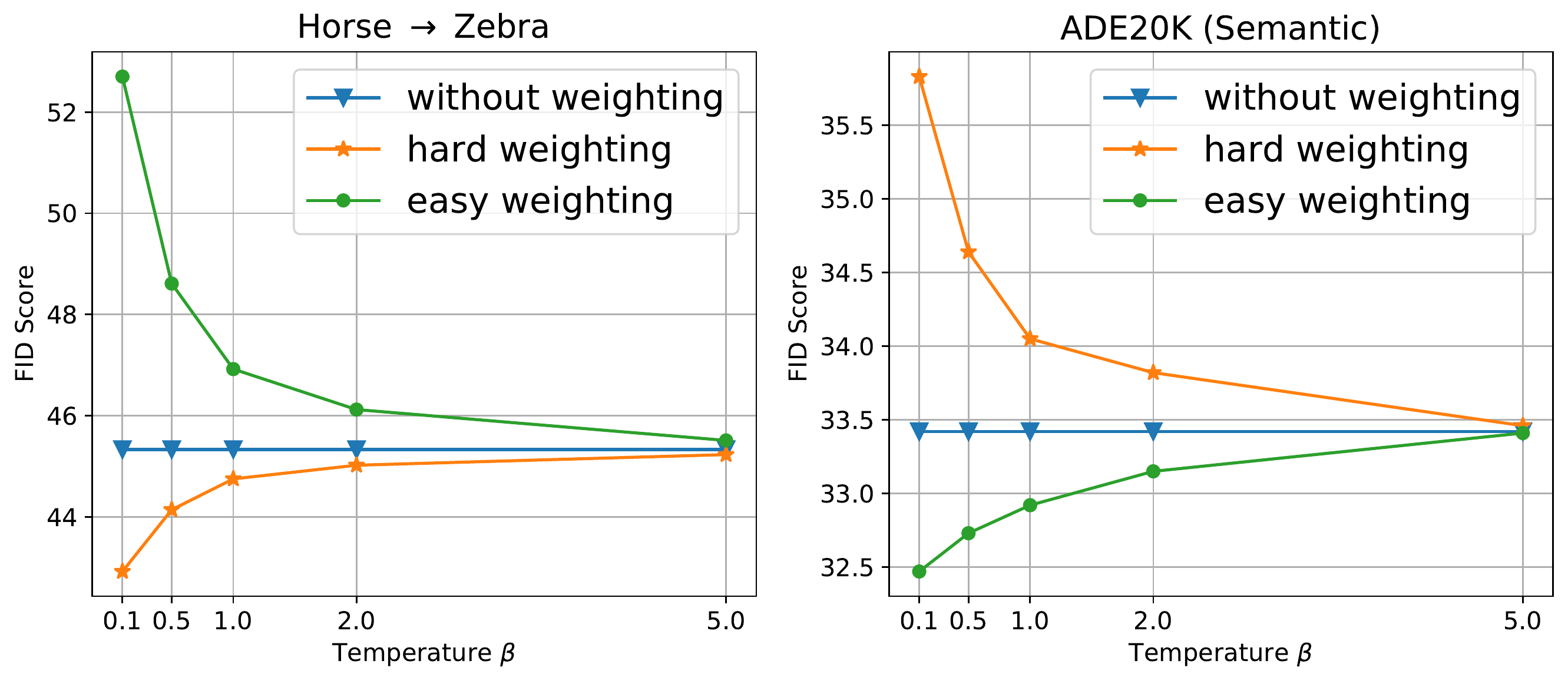}
\caption{
Image translation performance (FID) using different weighting strategies with varying temperature $\beta$.
The two graphs are the results of unpaired image translation (Horse $\rightarrow$ Zebra using CUT model \cite{park2020contrastive} with WeightNCE) and paired image translation (ADE20K using SPADE model \cite{park2020contrastive} with WeightNCE), respectively.
}
\label{im_weighting}
\end{figure}

\subsection{Weighted Contrastive Objective}
\label{sec_weightnce}

As each negative sample poses different similarity to anchor, the pushing force of each negative sample should be accordingly adjusted for better contrastive learning \cite{wang2019multi}.
To adjust the pushing force of a negative samples, a simple yet feasible approach is to adjust its weight in the contrastive objective.
According to Eq. (\ref{patchnce}), a higher weight of a negative pair (e.g., $e^{x_i\cdot y_j / \tau}$) indicates a higher importance in contrastive objective, i.e., enlarged pushing force for this negative pair.
Thus, the weighted version of Eq. (\ref{patchnce}) (denoted by \textbf{WeightNCE}) can be formulated as:
\begin{equation}
\label{weightnce}
-\sum_{i=1}^N \log
\frac{e^{x_i \cdot y_i / \tau}} 
{e^{x_i \cdot y_i / \tau} + Q (N-1) \sum_{\substack{j=1 \\ j\neq i}}^N w_{ij} \cdot e^{x_i \cdot y_j / \tau}},
\end{equation}
where $Q$ denotes the weight of negative terms ($Q=1$ by default) in the denominator, $w_{ij}$ ($j\neq i$) denotes the weight between sample $y_j$ and anchor $x_i$ and is subjected to $\sum_{\substack{j=1 \\ j\neq i}}^{N}w_{ij}=1, i\in [1, N]$.

The weighting strategy could essentially boil down to two categories: assigning higher weights to hard negative samples (referred as \textbf{hard weighting} $w_{ij}^+$) and assigning higher weights to easy negative samples (referred as \textbf{easy weighting} $w_{ij}^-$).
To determine the weighting strategy for unpaired and paired image translation, we illustrate the similarity histograms of positive and negative pairs in three middle layers (4th, 8th, 12th for unpaired image translation, $relu2\_2, relu3\_2, relu4\_2$ for paired image translation) after the contrastive learning is completed.
As shown in Fig. \ref{im_distribution}, for unpaired image translation, there is few overlap between the similarity histograms of positive and negative pairs after contrastive learning, which indicates that positive and negative pairs can be easily pushed apart.
In this end, the hard weighting strategy may help to boost the performance, as the model can focus on learning from more informative negative samples (hard negative samples) which has been proved to be beneficial for contrastive learning \cite{wang2019multi,robinson2020contrastive}.
However, for paired image translation, there is severe overlap for the similarity histogram of positive and negative pairs, which indicates many negative samples are hard to be distinguished from the positive samples.
In this case, hard weighting may not make for contrastive learning as naively using too hard negative samples may degrade the contribution of moderate ones, yielding worse representation \cite{jeon2021mining}.
It is reasonable to conjecture that easy weighting may contribute to the contrastive learning in this scenario by assigning lower weights to these hard negative samples which reduces their effects in the contrastive objective.

\begin{figure*}[t]
\centering
\includegraphics[width=1.0\linewidth]{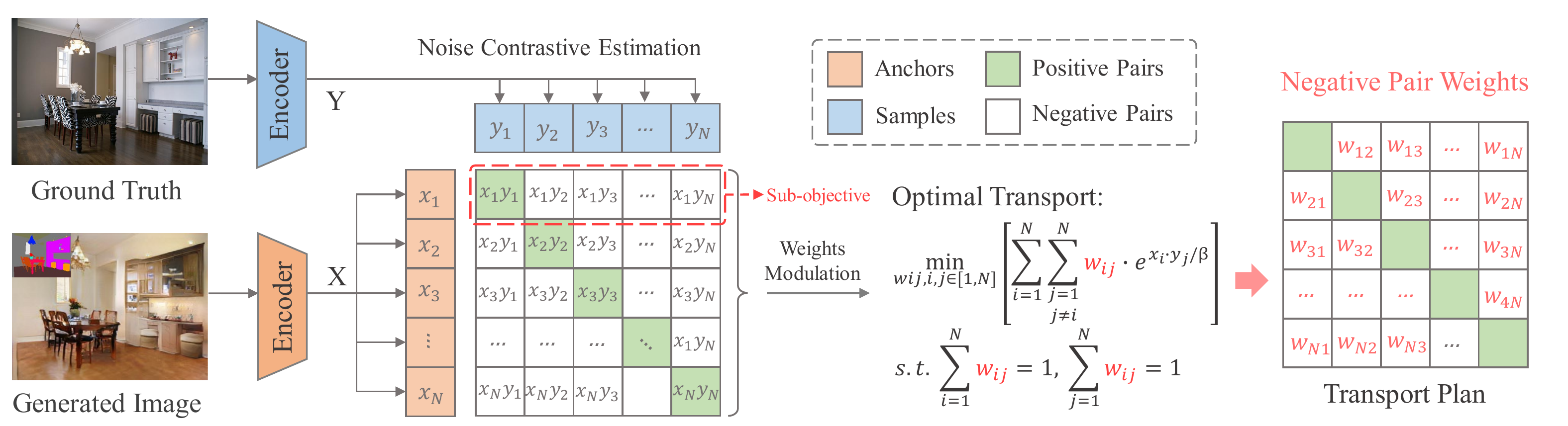}
\caption{
Framework of the proposed modulated contrast. 
There are multiple sub-objectives for the contrastive learning between feature set $X=[x_1, x_2, x_3, \cdots, x_N]$ and $Y=[y_1, y_2, y_3, \cdots, y_N]$.
To modulate the weights of negative pairs across multiple sub-objectives, optimal transport with a cost matrix $C$ (defined by $C_{ij}= e^{x_i\cdot y_j  / \beta}$ for unpaired translation,  $C_{ij}= e^{(1-x_i\cdot y_j) / \beta}$ for paired translation) is conducted between feature sets $X$ and $Y$ to minimize the total transport cost, yielding an optimal transport plan which serves as the weights of the corresponding negative pairs.
}
\label{im_stru}
\end{figure*}

We validate above conjecture by apply both hard weighting and easy weighting strategy to unpaired and paired image translation, respectively.
For the contrastive objective of a single patch,
hard weighting weights $w_{ij}^{+}$ and easy weighting weights $w_{ij}^{-}$ are determined with a positive and negative relation to the similarity between sample $y_j$ and anchor $x_i$ as below:
\begin{equation}
\label{weight}
w_{ij}^{+} = 
\frac{e^{(x_i \cdot y_j) / \beta}} 
{\sum_{j=1}^N e^{(x_i \cdot y_j) / \beta}}  \quad
w_{ij}^{-} = 
\frac{e^{(1 - x_i \cdot y_j) / \beta}} 
{\sum_{j=1}^N e^{(1 - x_i \cdot y_j) / \beta}}, 
\end{equation}
where $\beta$ denotes the weighting temperature parameter.
We take the value of temperature $\beta$ and the FID score of generated images as the abscissa and ordinate, respectively, as show in Fig. \ref{im_weighting}.
Treating the generation performance without weighting as the baseline, 
we can observe that the performance of unpaired image translation benefits from hard weighting strategy, and presents a positive correlation with the decreasing of $\beta$.
On the other hand, paired image translation performance benefits from easy weighting strategy, and also presents a positive correlation with the decreasing of $\beta$, which is consistent with our conjecture.
Despite some previous work \cite{wang2019multi,robinson2020contrastive} proves the effectiveness of hard negative samples for contrastive learning, we would clarify that the bad effect of excessively hard samples have overwhelmed their positive effect in the case of paired image translation.
% The dilemma is caused by the inherent defect of unsupervised contrastive loss that it pushes all different instances ignoring their semantical relation.

% In above experiments, the weighting strategy in Eq. (\ref{weight}) are applied to each contrastive sub-objective separately.
% However, as all contrastive sub-objectives are correlated to form the final objective as in Eq. \ref{weightnce}, several, x
% weighting each sub-objective separately without overall coordination tends to be sub-optimal as there may be some conflicts for the weighting strategies of different sub-objectives.
In the above experiments, the weighting strategy in Eq. (\ref{weight}) is applied to each contrastive sub-objective separately.
However, all contrastive sub-objectives are contributing to the final objective as in Eq. \ref{weightnce}. Weighting each sub-objective independently without overall coordination may result in conflicts between different sub-objectives, and thus tends to be sub-optimal for the final objective.

\subsection{Modulated Contrastive Objective}
\label{sec_monce}

As we are exploring \textbf{re-weighing} strategies, the total weight associated with a feature ($x_i$ or $y_j$) is expected to be constant, thus yielding below constraints:
\begin{equation}
\label{monce_constraint}
\sum_{i=1}^{N}w_{ij}=1,  \quad  \sum_{j=1}^{N}w_{ij}=1, \quad i,j\in[1,N].
\end{equation} 
Considering the contrastive objective as illustrated in Fig. \ref{im_stru}, a feature $y_j$ serves as negative sample for multiple sub-objectives. As the total weights associated with $y_j$ is constant (i.e., $\sum_{i=1}^{N}w_{ij}=1$), there may be conflicts for the weighting strategies of $y_j$ in different sub-objectives, e.g., several sub-objectives all expect a higher weight for $y_j$ while the total weights of $y_j$ is constrained.
% $x_i, i \in [1, N], i \neq j$ with weights $w_{ij}$.
Therefor, we aims to modulate the assignment of weights $w_{ij}$ ($i,j \in [1,N], i \neq j$) across multiple sub-objectives with the constraint of constant total weight.

% To find the optimal weights assignments for all negative pairs,
% re-weight all negative pairs collaborately.
Targeting to modulate the weights assignment for all negative pairs,
a weight modulation goal shared across all contrastive sub-objectives should be determined.
We take easy weighting strategy as an example to derive the final weight modulation goal.
By assigning higher weights to negative pairs with low similarity, the easy weighting strategy for a contrastive sub-objective in Eq. (\ref{weightnce}) is equivalent to reducing the negative term $\sum_{\substack{j=1 \\ j\neq i}}^N w_{ij} \cdot e^{x_i \cdot y_j / \tau}$.
As the contrastive objectives of all image patches are summed to form the final objective, the shared modulation goal across multiple contrastive objectives can be regarded as reducing the total loss of negatives terms.
To derive the expression mathematically, the objective of the modulation goal is formulated as \textit{minimizing} the total loss of negative terms with regarding to $w_{ij},i,j\in[1,N]$:
\begin{equation}
\label{monce}
\underset{w_{ij}, i,j\in[1,N]}{\min} 
\Bigg [ 
\sum_{i=1}^N 
\sum_{\substack{j=1 \\ j\neq i}}^N w_{ij} \cdot e^{x_i \cdot y_j / \tau}
\Bigg ] .
\end{equation}
The formation of Eq. (\ref{monce}) with constraint in Eq. (\ref{monce_constraint}) can be regarded as an optimal transport (OT) \cite{peyre2019computational} problem between $[x_1, x_2, \cdots , x_N]$ and $[y_1, y_2, \cdots , y_N]$ with a cost matrix $C$ defined by $C_{ij}= e^{x_i\cdot y_j  / \beta}$ for $i\neq j$ and $C_{ij}=\inf$ for $i=j$.
Similar to weighting temperature in Eq. (\ref{weight}),
$\beta$ in the cost matrix $C$ serves as a cost temperature that indicates the smoothness of the optimal transport.
A smaller $\beta$ tends to assign higher weights for small cost entries $C_{ij}$ and a large $\beta$ tends to assign equal weights for all cost entries.
Detailed parameter study of $\beta$ can be found in the experiment part.

The optimal transport aims to retrieve a transport plan $T$ which minimizes the total transport cost as formulated below:
\begin{equation}
\label{sinkhorn}
\mathop{\min}\limits_{T} 
 \langle C, T \rangle ,
 \quad  s.t.  \ \  (T \vec{1}) = 1, \ (T^\top \vec{1}) = 1, 
\end{equation}
where $\langle C, T \rangle$ denotes the inner product of $C$ and $T$.
Thus, solving the transport plan $T$ is equivalent to solve the weight parameters as $w_{ij} = T_{ij}$.
The Sinkhorn algorithm \cite{cuturi2013sinkhorn} can be applied to Eq. (\ref{sinkhorn}) for approximating optimal transport solution, yielding the desired optimal transport plan $T$.
With the derived transport plan matrix $T$ as the weights of negative pairs, the modulated objective for easy weighting strategy is accordingly determined.
For hard weighting strategy, the modulated objective can be derived similarly, just redefining the cost matrix $C$ in Eq. \ref{sinkhorn} as $C_{ij}=e^{(1-x_i y_j) / \beta}$ for $i \neq j$ and $C_{ij}=\inf$ for $i=j$.

\begin{figure*}[t]
\centering
\includegraphics[width=1.0\linewidth]{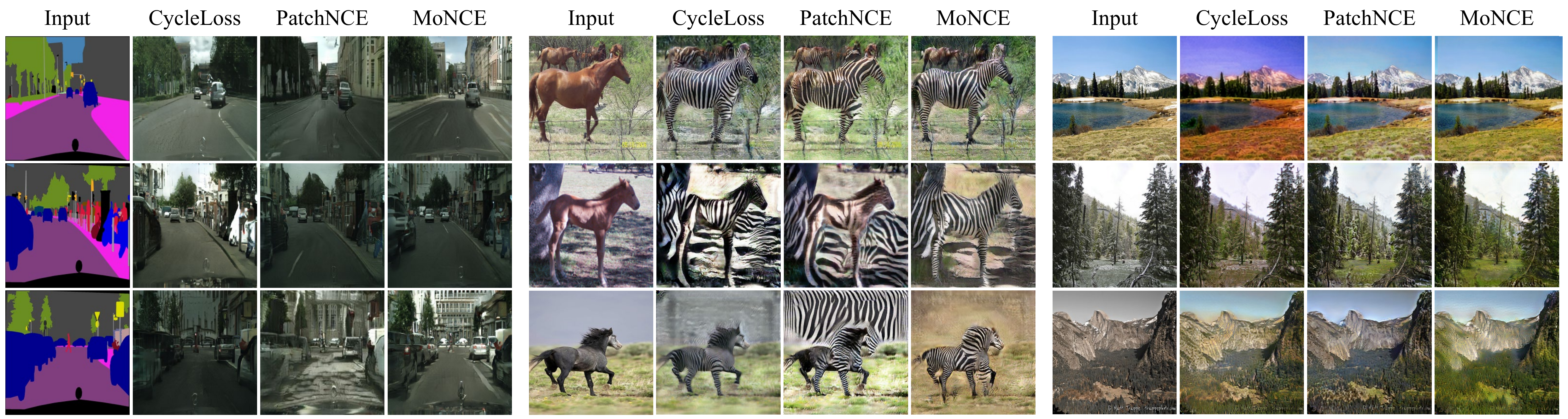}
\caption{
Qualitative comparison of different losses for unpaired image translation tasks including Cityscapes (Semantic $\rightarrow$ Image), Horse $\rightarrow$ Zebra, Winter $\rightarrow$ Summer. The structure of CUT \cite{park2020contrastive} is employed for the translation.
}
\label{im_cut_samples}
\end{figure*}

\renewcommand\arraystretch{1.2}
\begin{table*}[t]
\small 
\renewcommand\tabcolsep{10pt}
\centering 
\begin{tabular}{l||cccc||cc||cc} 
\hline
& 
\multicolumn{4}{c||}{\textbf{Cityscapes (Semantic $\rightarrow$ Image)}} & 
\multicolumn{2}{c||}{\textbf{Horse $\rightarrow$ Zebra}} &
\multicolumn{2}{c}{\textbf{Winter $\rightarrow$ Summer}}
\\
\cline{2-9}
\multirow{-2}{*}{\textbf{CUT \cite{park2020contrastive}}} 
& FID $\downarrow$ & mAP $\uparrow$ & pixAcc $\uparrow$ & classAcc $\uparrow$ 
& FID $\downarrow$ & SWD  $\downarrow$ 
& FID $\downarrow$ & SWD $\downarrow$ 
\\\hline

\textbf{Baseline (GAN Loss)} & 139.9 & 9.705 & 23.44 & 14.17     & 129.8 & 74.85     & 136.2  & 47.80                  \\

\textbf{+Cycle Loss \cite{zhu2017unpaired}}     &  75.97 & 20.53 & 55.87 & 25.23     & 76.37 & 50.54       & 86.14 & 38.79              \\

\textbf{+PatchNCE\cite{park2020contrastive}} & 57.16 &  24.29 & 78.22 & 30.67      & 45.33 & 32.02        & 80.25  & 36.92               \\

\rowcolor{mygray} \textbf{+WeightNCE}     & 55.94 &  24.98  & 77.92 & 31.96      & 42.92 & 31.58       & 79.32  & 36.39     \\

\rowcolor{mygray} \textbf{+MoNCE}
& \textbf{54.67} & \textbf{25.61} & \textbf{78.41} & \textbf{33.02}
& \textbf{41.86} & \textbf{30.80}
& \textbf{78.18} & \textbf{35.95}
  \\
  \hline
\end{tabular}

\caption{
Unpaired image translation performance on different tasks with CUT \cite{park2020contrastive} as the model structure.
}

\label{tab_cut}
\end{table*}

\renewcommand\arraystretch{1.4}
\begin{table}[t]
\footnotesize
\renewcommand\tabcolsep{7pt}
\centering 
\begin{tabular}{l|cc|cc} 
\hline
& 
\multicolumn{2}{c|}{\textbf{Horse$\rightarrow$Zebra}} &
\multicolumn{2}{c}{\textbf{Winter$\rightarrow$Summer}}
\\
\cline{2-5}
\multirow{-2}{*}{\textbf{F/LSeSim \cite{zheng2021spatially}}} 
& FID $\downarrow$ & SWD  $\downarrow$ 
& FID $\downarrow$ & SWD $\downarrow$ 
\\\hline
\hline

\textbf{Random SeSim \cite{kolkin2019style}}      & 72.18 & 48.85     & 125.1  & 57.48                  \\

\textbf{FSeSim}         & 43.26 & 36.77       & 79.14 & 35.79              \\
\textbf{LSeSim+PatchNCE}       & 40.12 & 34.77       & 78.30 &  34.47         \\
\rowcolor{mygray} \textbf{LSeSim+WeightNCE}       & 38.67 & 32.59       & 76.98 &  33.89         \\
\rowcolor{mygray} \textbf{LSeSim+MoNCE}     & \textbf{37.21} & \textbf{32.12}   & \textbf{76.04} & \textbf{33.10}         \\\hline
\end{tabular}
\caption{
Unpaired image translation with F/LSeSim \cite{zheng2021spatially} as the model structure.
}
\label{tab_sc}
\end{table}

\section{Experiments}

\subsection{Experimental Settings}

\textbf{Datasets:}
For unpaired image translation, we conducted experiments on Cityscapes, Horse $\rightarrow$ Zebra, and Winter $\rightarrow$ Summer. For paired image translation, we conducted experiments on ADE20K, CelebA-HQ, and DeepFashion.

\noindent
$\bullet$ Cityscapes \cite{cordts2016cityscapes} contains 2,975 training and 500 validation images captured on street. 
We conduct unpaired semantic-to-image translation on this dataset.

\noindent
$\bullet$ Horse $\rightarrow$ Zebra \cite{zhu2017unpaired} collects 1187 horse images and 1474 zebra images from ImageNet \cite{deng2009imagenet} for training and validation.

\noindent
$\bullet$ Winter $\rightarrow$ Summer \cite{zhu2017unpaired} contains 1,200 winter images and 1,540 summer images for training and validation.

\noindent
$\bullet$ ADE20k \cite{zhou2017ade20k} consists of 20k training images with 150-class segmentation masks. 
We conduct image generation by using its semantic segmentation as conditional inputs.

\noindent
$\bullet$ CelebA-HQ \cite{liu2015celebahq} consists of 30,000 face images. We use its semantic map and edge maps for conditional generation.

\noindent
$\bullet$ DeepFashion \cite{liu2016deepfashion} contains 52,712 person images.
We use its keypoints as conditional inputs in experiments.

\begin{figure*}[t]
\centering
\includegraphics[width=1.0\linewidth]{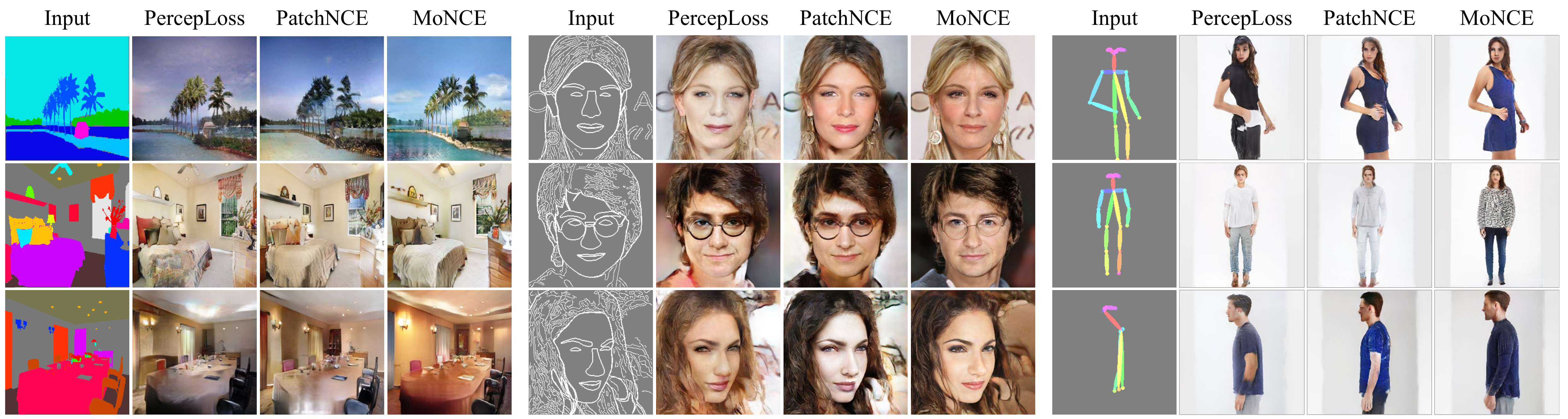}
\caption{
Qualitative comparison of different losses for paired image translation tasks including ADE20K (Semantic), CelebA-HQ (Edge), and DeepFashion (Keypoint). The structure of SPADE \cite{park2019spade} is employed for the translation.
}
\label{im_spade_samples}
\end{figure*}

\renewcommand\arraystretch{1.2}
\begin{table*}[ht]
\small 
\renewcommand\tabcolsep{5.5pt}
\centering 
\begin{tabular}{l||ccc||cc||cc||cc} 
\hline
& 
\multicolumn{3}{c||}{\textbf{ADE20K (Semantic)}} & 
\multicolumn{2}{c||}{\textbf{CelebA-HQ (Semantic)}} &
\multicolumn{2}{c||}{\textbf{CelebA-HQ (Edge)}} &
\multicolumn{2}{c}{\textbf{DeepFashion (Keypoint)}}
\\
\cline{2-10}
\multirow{-2}{*}{\textbf{SPADE \cite{park2019spade}}} 
& FID $\downarrow$ & mIoU $\uparrow$ & Acc $\uparrow$ 
& FID $\downarrow$ & SWD $\downarrow$ 
& FID $\downarrow$ & SWD $\downarrow$ 
& FID $\downarrow$ & SWD $\downarrow$  \\\hline 

\textbf{Baseline (GAN Loss)} 
& 87.32 & 31.32 & 76.79        & 86.91 & 25.93      & 84.04 & 27.35        & \textbf{28.57} & 22.18      \\

\textbf{+PercepLoss \cite{johnson2016perceptual}}     
&  33.68 & 42.23 & 81.96            & 36.54 & \textbf{17.28}       & 31.53 & 18.25     & 35.74 & 24.03     \\

\textbf{+PatchNCE\cite{park2020contrastive}} 
& 33.42 & 44.91 & 81.92         & 33.38 & 21.90         & 30.81 &  23.14      & 38.04 & 23.53   \\

\rowcolor{mygray}  \textbf{+WeightNCE}     
& 32.47 &  45.69  & 83.52             & 32.15 & 18.86       & 30.49 & 21.51          & 36.83 & 22.98   \\

\rowcolor{mygray} \textbf{+MoNCE}
& \textbf{31.62} & \textbf{46.30} & \textbf{84.29}
& \textbf{30.01} & 17.39 
& \textbf{29.75} & \textbf{18.11}
& 33.96 & \textbf{21.58} 
  \\\hline
\end{tabular}
\caption{
Paired image translation with different types of conditional input. The model structure of SPADE \cite{park2019spade} is employed to compare the performance of different losses.
}
\label{tab_spade}
\end{table*}

\textbf{Evaluation Metrics:}
Several evaluation metrics are adopted in our experiment to assess image translation performance.
\textit{Fr{\'e}chet Inception Score (FID)} \cite{fid} and sliced \textit{Wasserstein distance (SWD)} \cite{swd} are adopted to measure distribution discrepancy and statistical distances of low level patches between translated images and real images, respectively.
For semantic image translation tasks, we employ pre-trained segmentation model to evaluate the segmentation accuracy, e.g., mean average precision (mAP) and pixel accuracy (Acc).
% More details of the pre-trained segmentation models are provided in the supplementary material.

\textbf{Implementation Details:}
All experiments are conducted with an image resolution of 256$\times$256.
For contrastive learning setting, we keep the same with CUT \cite{park2020contrastive}, e.g., 256 negative samples, temperature parameter $\tau$ = 0.07.
The default temperatures $\beta$ and weight term weight $Q$ in WeightNCE and MoNCE are 0.1 and 1, respectively, for all tasks.
% The negative term weight $Q$ is adaptively adjusted to ensure that the value of negative term are in the same level.
% For unpaired image translation, the experiments are conducted on one V100 GPU with a batch size of 12.
% For paired image translation, the experiments are conducted on four V100 GPUs with a batch size of 80.
We re-train all compared methods following above setting to ensure fair comparison.
% More detailed training setting can be found in the supplementary material.
% (learning rate, optimizer, etc.)

\subsection{Unpaired Image Translation}

We evaluate our proposed loss on the classical unpaired image translation task.
% that aims for the preservation of image contents.
We first adopt the model structure of CUT \cite{park2020contrastive} to conduct comparison between CycleLoss \cite{zhu2017unpaired}, PatchNCE loss \cite{park2020contrastive}, and our proposed WeightNCE and MoNCE.
Complying with the discussion in the Sec. \ref{sec_weightnce}, the weighting strategy of assigning higher weights to hard negative samples is adopted for unpaired image translation.
As shown in Table \ref{tab_cut}, the model with GAN Loss only is used as the Baseline. The four different losses are further included into the Baseline, respectively, for comparisons.
We can observe that the proposed WeightNCE and MoNCE both outperform the CycleLoss and PatchNCE consistently in all compared unpaired translation tasks.
With an overall weight modulation across multiple contrastive objectives, the proposed MoNCE outperforms WeightNCE across all evaluation metrics.
Fig. \ref{im_cut_samples} shows qualitative comparisons on unpaired image translation. All compared methods adopt the same structure with CUT and the only variation comes from different losses. 
% It can be observed that the MoNCE translated images preserves better contents and structures clearly and consistently across all tasks.

Besides CUT model, we also compare the four losses with the F/LSeSim \cite{zheng2021spatially} model,
which exploits the spatial patterns of self-similarity to preserve image structures in unpaired image translation.
The content loss \cite{kolkin2019style} using random sampled features for computing self-similarity is selected as the baseline (Random SeSim).
F/LSeSim could employ a pre-trained VGG-16 \cite{simonyan2014very} (namely FSeSim) or a PatchNCE (namely LSeSim) to learn spatially correlative maps.
Here, we replace the PatchNCE with our WeightNCE and MoNCE to conduct the comparison.
As shown in Table \ref{tab_sc}, the learnable self-similarity setting (LSeSim+PatchNCE) outperforms the fixed self-similarity setting with pre-trained VGG-16 \cite{simonyan2014very}.
Consistent with the results in CUT, replacing the PatchNCE with our WeightNCE and MoNCE also bring notable improvement in translation quality.

\begin{figure*}[t]
\centering
\includegraphics[width=1.0\linewidth]{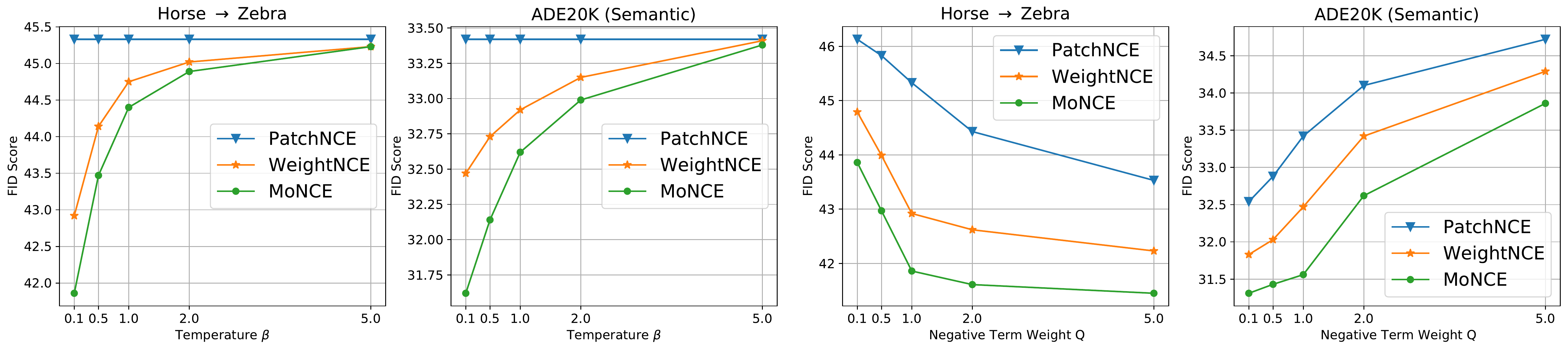}
\caption{
The effect of varying temperature parameter $\beta$ and negative term weight $Q$ on unpaired image translation (Horse $\rightarrow$ Zebra) and paired image translation (ADE20K (Semantic)).
}
\label{im_ablation}
\end{figure*}

\subsection{Paired Image Translation}

For paired image translation, we adopt the structure of SPADE \cite{park2019spade} to perform the comparison between PercepLoss \cite{johnson2016perceptual}, PatchNCE \cite{park2020contrastive}, and our proposed WeightNCE and MoNCE.
The SPADE model with GAN loss only is selected as the baseline.
Then the Baseline is combined with different losses to perform the comparisons.
As shown in Table \ref{tab_spade}, PatchNCE with pre-trained VGG-19 for feature extraction could rival the well-known PercepLoss across all generation tasks in terms of generation quality.
Considering the performance of vanilla PatchNCE in unpaired and paired image translation, contrastive learning has good potential to serve as a versatile metric for measuring image similarity.
Besides, we can observe the MoNCE is advantageous to WeightNCE and both of them outperform the PatchNCE consistently, which verify the effectiveness of our weighting strategies and modulation mechanism.
% (designed based on point-wise deviation)

Fig. \ref{im_spade_samples} shows qualitative experiments with SPADE with different losses. 
We can see that the images translated with PatchNCE tends to present less artifacts compared that with PercepLoss, as contrastive learning aims to maximize the mutual information of corresponding images instead of naively minimizing the point-wise absolute deviation.
With an overall modulation of easy weighting stategies,
% that assigns less weights to hard negative samples. 
our MoNCE outperforms PatchNCE clearly with more fine details in generated images.
% We conjecture that some negative samples are very similar to positive samples in PatctNCE and pushing apart these very hard negative samples undesirably degrades the contrastive learning.

\renewcommand\arraystretch{1.1}
\begin{table}[t]
\small
\renewcommand\tabcolsep{5pt}
\centering 
\begin{tabular}{p{1.4cm}<{\centering} p{1.6cm}<{\centering}  p{1.1cm}<{\centering}  p{1.1cm}<{\centering} p{1.1cm}<{\centering} }
\hline

\multicolumn{3}{c}{\textbf{MoNCE Variants}} &  \multicolumn{2}{c}{\textbf{Performance}} \\

\cmidrule(r){1-3} \cmidrule(r){4-5} 
Bidirectional & Pre-trained & Frozen & FID $\downarrow$ & mIoU $\uparrow$ \\
\hline
\textcolor{green}{\checkmark} & \textcolor{green}{\checkmark}  & \textcolor{green}{\checkmark}  & 31.31  & \textbf{46.71}   \\
\textcolor{green}{\checkmark}  & \textcolor{green}{\checkmark}  &  \textcolor{red}{\xmark}   & \textbf{30.94}  & 44.89    \\
\textcolor{green}{\checkmark}  & \textcolor{red}{\xmark}  &  \textcolor{red}{\xmark}   & 40.47  & 38.22    \\
\textcolor{red}{\xmark}  & \textcolor{red}{\xmark}  &  \textcolor{red}{\xmark}   & 42.86  & 36.16    \\
\textcolor{red}{\xmark}  & \textcolor{green}{\checkmark}  &  \textcolor{red}{\xmark}   & 31.37  & 45.44    \\
\rowcolor{mygray}  \textcolor{red}{\xmark}  & \textcolor{green}{\checkmark}  &  \textcolor{green}{\checkmark}   & 31.62  & 46.30    \\
\hline
\end{tabular}
\caption
{
Ablation study of MoNCE variants on paired image translation (ADE20K).
The configuration at the grey row is the default setting of MoNCE.
}
\label{tab_variants}
\end{table}

\subsection{Discussion}

We conduct experiments on unpaired image translation (Horse $\rightarrow$ Zebra) and paired image translation (ADE20K (Semantic)) to examine the effect of the cost temperature $\beta$ in Eq. (\ref{sinkhorn}).
As show in Fig. \ref{im_ablation}, 
the generation performance (FID score) of unpaired and paired image translation improves consistently while decreasing the temperature $\beta$.
However, we find the model training tends to be unstable and even fail with small temperature $\beta$, e.g., 0.01.
We also ablate the effect of the negative terms weight $Q$ in Eq. (\ref{weightnce}).
As shown in Fig. \ref{im_ablation}, the performance of unpaired image translation and paired image translation presents positive correlation and negative correlation with the increasing of negative term weight $Q$, respectively.
Although the FID is improved with a larger $Q$, we observe that the content preservation performance is actually degraded for unpaired image translation.
Based on above observation, we set the temperature $\beta$ as 0.1 and the negative term weight $Q$ as 1 by default.
% Please refer to the supplementary material for more detailed analysis.

We also explore several variants of contrastive learning on paired image translation (ADE20K), including without pre-trained VGG-19 network, unfrozen pre-trained VGG-19 network, and bidirectional designing of contrastive learning (including the contrastive objective with ground truth patches as anchors) introduced in \cite{andonian2021contrastive}.
As shown in Table \ref{tab_variants}, 
learning the feature extractor from scratch without pre-training tends to impair the generation performance drastically.
% the generation performance is much lower than the standard MoNCE when learning the feature extractor from scratch without pre-training.
Including bidiretional design to the proposed MoNCE improve the generation performance slightly.
Unfreezing the pre-trained VGG-19 network improves the FID, while it hurts the mAP score.

\section{Conclusion}

We have formulated contrastive learning as a versatile metric for various image translation tasks, which is on par with the prevailing losses designed in corresponding tasks.
With a target to re-weighting negative pairs for performance gain, we explore and establish the weighting strategies for unpaired and paired image translation according to the similarity distribution of positive and negative pairs.
To modulate the re-weighting of all negative pairs associated with the full image, we further derive a MoNCE which employs optimal transport to retrieve the optimal weights for negative pairs across multiple contrastive objectives.
Our thorough and extensive analysis of negative pair weighting strategies lays a sound foundation for the exploration of contrastive learning in image generation.

\section{Acknowledgement}
This study is supported under the RIE2020 Industry Alignment Fund – Industry Collaboration Projects (IAF-ICP) Funding Initiative, as well as cash and in-kind contribution from the industry partner(s).

%%%%%%%%% REFERENCES
{\small
\bibliographystyle{ieee_fullname}
\bibliography{egbib}

\begin{thebibliography}{10}\itemsep=-1pt

\bibitem{amodio2019travelgan}
Matthew Amodio and Smita Krishnaswamy.
\newblock Travelgan: Image-to-image translation by transformation vector
  learning.
\newblock In {\em Proceedings of the IEEE/CVF Conference on Computer Vision and
  Pattern Recognition}, pages 8983--8992, 2019.

\bibitem{andonian2021contrastive}
Alex Andonian, Taesung Park, Bryan Russell, Phillip Isola, Jun-Yan Zhu, and
  Richard Zhang.
\newblock Contrastive feature loss for image prediction.
\newblock In {\em Proceedings of the IEEE/CVF International Conference on
  Computer Vision}, pages 1934--1943, 2021.

\bibitem{benaim2017one}
Sagie Benaim and Lior Wolf.
\newblock One-sided unsupervised domain mapping.
\newblock {\em Advances in Neural Information Processing Systems}, 30, 2017.

\bibitem{chen2021large}
Shuo Chen, Gang Niu, Chen Gong, Jun Li, Jian Yang, and Masashi Sugiyama.
\newblock Large-margin contrastive learning with distance polarization
  regularizer.
\newblock In {\em International Conference on Machine Learning}, pages
  1673--1683. PMLR, 2021.

\bibitem{chen2020simple}
Ting Chen, Simon Kornblith, Mohammad Norouzi, and Geoffrey Hinton.
\newblock A simple framework for contrastive learning of visual
  representations.
\newblock In {\em International conference on machine learning}, pages
  1597--1607. PMLR, 2020.

\bibitem{chuang2020debiased}
Ching-Yao Chuang, Joshua Robinson, Lin Yen-Chen, Antonio Torralba, and Stefanie
  Jegelka.
\newblock Debiased contrastive learning.
\newblock {\em arXiv preprint arXiv:2007.00224}, 2020.

\bibitem{cordts2016cityscapes}
Marius Cordts, Mohamed Omran, Sebastian Ramos, Timo Rehfeld, Markus Enzweiler,
  Rodrigo Benenson, Uwe Franke, Stefan Roth, and Bernt Schiele.
\newblock The cityscapes dataset for semantic urban scene understanding.
\newblock In {\em Proceedings of the IEEE conference on computer vision and
  pattern recognition}, pages 3213--3223, 2016.

\bibitem{cuturi2013sinkhorn}
Marco Cuturi.
\newblock Sinkhorn distances: Lightspeed computation of optimal transport.
\newblock In {\em Advances in neural information processing systems}, pages
  2292--2300, 2013.

\bibitem{deng2009imagenet}
Jia Deng, Wei Dong, Richard Socher, Li-Jia Li, Kai Li, and Li Fei-Fei.
\newblock Imagenet: A large-scale hierarchical image database.
\newblock In {\em 2009 IEEE conference on computer vision and pattern
  recognition}, pages 248--255. Ieee, 2009.

\bibitem{deng2020disentangled}
Yu Deng, Jiaolong Yang, Dong Chen, Fang Wen, and Xin Tong.
\newblock Disentangled and controllable face image generation via 3d
  imitative-contrastive learning.
\newblock In {\em Proceedings of the IEEE/CVF Conference on Computer Vision and
  Pattern Recognition}, pages 5154--5163, 2020.

\bibitem{dosovitskiy2016generating}
Alexey Dosovitskiy and Thomas Brox.
\newblock Generating images with perceptual similarity metrics based on deep
  networks.
\newblock {\em Advances in neural information processing systems}, 29:658--666,
  2016.

\bibitem{fu2019geometry}
Huan Fu, Mingming Gong, Chaohui Wang, Kayhan Batmanghelich, Kun Zhang, and
  Dacheng Tao.
\newblock Geometry-consistent generative adversarial networks for one-sided
  unsupervised domain mapping.
\newblock In {\em Proceedings of the IEEE/CVF Conference on Computer Vision and
  Pattern Recognition}, pages 2427--2436, 2019.

\bibitem{gatys2016image}
Leon~A Gatys, Alexander~S Ecker, and Matthias Bethge.
\newblock Image style transfer using convolutional neural networks.
\newblock In {\em Proceedings of the IEEE conference on computer vision and
  pattern recognition}, pages 2414--2423, 2016.

\bibitem{he2020momentum}
Kaiming He, Haoqi Fan, Yuxin Wu, Saining Xie, and Ross Girshick.
\newblock Momentum contrast for unsupervised visual representation learning.
\newblock In {\em Proceedings of the IEEE/CVF Conference on Computer Vision and
  Pattern Recognition}, pages 9729--9738, 2020.

\bibitem{fid}
Martin Heusel, Hubert Ramsauer, Thomas Unterthiner, Bernhard Nessler, and Sepp
  Hochreiter.
\newblock Gans trained by a two time-scale update rule converge to a local nash
  equilibrium.
\newblock In {\em Advances in neural information processing systems}, pages
  6626--6637, 2017.

\bibitem{jeon2021mining}
Sangryul Jeon, Dongbo Min, Seungryong Kim, and Kwanghoon Sohn.
\newblock Mining better samples for contrastive learning of temporal
  correspondence.
\newblock In {\em Proceedings of the IEEE/CVF Conference on Computer Vision and
  Pattern Recognition}, pages 1034--1044, 2021.

\bibitem{johnson2016perceptual}
Justin Johnson, Alexandre Alahi, and Li Fei-Fei.
\newblock Perceptual losses for real-time style transfer and super-resolution.
\newblock In {\em European conference on computer vision}, pages 694--711.
  Springer, 2016.

\bibitem{kang2020ContraGAN}
Minguk Kang and Jaesik Park.
\newblock {ContraGAN: Contrastive Learning for Conditional Image Generation}.
\newblock In {\em Conference on Neural Information Processing Systems
  (NeurIPS)}, 2020.

\bibitem{swd}
Tero Karras, Timo Aila, Samuli Laine, and Jaakko Lehtinen.
\newblock Progressive growing of gans for improved quality, stability, and
  variation.
\newblock {\em arXiv preprint arXiv:1710.10196}, 2017.

\bibitem{kolkin2019style}
Nicholas Kolkin, Jason Salavon, and Gregory Shakhnarovich.
\newblock Style transfer by relaxed optimal transport and self-similarity.
\newblock In {\em Proceedings of the IEEE/CVF Conference on Computer Vision and
  Pattern Recognition}, pages 10051--10060, 2019.

\bibitem{liu2016deepfashion}
Ziwei Liu, Ping Luo, Shi Qiu, Xiaogang Wang, and Xiaoou Tang.
\newblock Deepfashion: Powering robust clothes recognition and retrieval with
  rich annotations.
\newblock In {\em Proceedings of the IEEE conference on computer vision and
  pattern recognition}, pages 1096--1104, 2016.

\bibitem{liu2015celebahq}
Ziwei Liu, Ping Luo, Xiaogang Wang, and Xiaoou Tang.
\newblock Deep learning face attributes in the wild.
\newblock In {\em Proceedings of the IEEE international conference on computer
  vision}, pages 3730--3738, 2015.

\bibitem{mechrez2018maintaining}
Roey Mechrez, Itamar Talmi, Firas Shama, and Lihi Zelnik-Manor.
\newblock Maintaining natural image statistics with the contextual loss.
\newblock In {\em Asian Conference on Computer Vision}, pages 427--443.
  Springer, 2018.

\bibitem{mechrez2018contextual}
Roey Mechrez, Itamar Talmi, and Lihi Zelnik-Manor.
\newblock The contextual loss for image transformation with non-aligned data.
\newblock In {\em Proceedings of the European Conference on Computer Vision
  (ECCV)}, pages 768--783, 2018.

\bibitem{oord2018representation}
Aaron van~den Oord, Yazhe Li, and Oriol Vinyals.
\newblock Representation learning with contrastive predictive coding.
\newblock {\em arXiv preprint arXiv:1807.03748}, 2018.

\bibitem{park2020contrastive}
Taesung Park, Alexei~A Efros, Richard Zhang, and Jun-Yan Zhu.
\newblock Contrastive learning for unpaired image-to-image translation.
\newblock In {\em European Conference on Computer Vision}, pages 319--345.
  Springer, 2020.

\bibitem{park2019spade}
Taesung Park, Ming-Yu Liu, Ting-Chun Wang, and Jun-Yan Zhu.
\newblock Semantic image synthesis with spatially-adaptive normalization.
\newblock In {\em Proceedings of the IEEE Conference on Computer Vision and
  Pattern Recognition}, pages 2337--2346, 2019.

\bibitem{peyre2019computational}
Gabriel Peyr{\'e}, Marco Cuturi, et~al.
\newblock Computational optimal transport: With applications to data science.
\newblock {\em Foundations and Trends{\textregistered} in Machine Learning},
  11(5-6):355--607, 2019.

\bibitem{robinson2020contrastive}
Joshua Robinson, Ching-Yao Chuang, Suvrit Sra, and Stefanie Jegelka.
\newblock Contrastive learning with hard negative samples.
\newblock {\em arXiv preprint arXiv:2010.04592}, 2020.

\bibitem{shrivastava2017learning}
Ashish Shrivastava, Tomas Pfister, Oncel Tuzel, Joshua Susskind, Wenda Wang,
  and Russell Webb.
\newblock Learning from simulated and unsupervised images through adversarial
  training.
\newblock In {\em Proceedings of the IEEE conference on computer vision and
  pattern recognition}, pages 2107--2116, 2017.

\bibitem{simonyan2014very}
Karen Simonyan and Andrew Zisserman.
\newblock Very deep convolutional networks for large-scale image recognition.
\newblock {\em arXiv preprint arXiv:1409.1556}, 2014.

\bibitem{taigman2016unsupervised}
Yaniv Taigman, Adam Polyak, and Lior Wolf.
\newblock Unsupervised cross-domain image generation.
\newblock {\em arXiv preprint arXiv:1611.02200}, 2016.

\bibitem{ulyanov2017improved}
Dmitry Ulyanov, Andrea Vedaldi, and Victor Lempitsky.
\newblock Improved texture networks: Maximizing quality and diversity in
  feed-forward stylization and texture synthesis.
\newblock In {\em Proceedings of the IEEE Conference on Computer Vision and
  Pattern Recognition}, pages 6924--6932, 2017.

\bibitem{wang2021instance}
Weilun Wang, Wengang Zhou, Jianmin Bao, Dong Chen, and Houqiang Li.
\newblock Instance-wise hard negative example generation for contrastive
  learning in unpaired image-to-image translation.
\newblock In {\em Proceedings of the IEEE/CVF International Conference on
  Computer Vision}, pages 14020--14029, 2021.

\bibitem{wang2019multi}
Xun Wang, Xintong Han, Weilin Huang, Dengke Dong, and Matthew~R Scott.
\newblock Multi-similarity loss with general pair weighting for deep metric
  learning.
\newblock In {\em Proceedings of the IEEE/CVF Conference on Computer Vision and
  Pattern Recognition}, pages 5022--5030, 2019.

\bibitem{wang2004image}
Zhou Wang, Alan~C Bovik, Hamid~R Sheikh, and Eero~P Simoncelli.
\newblock Image quality assessment: from error visibility to structural
  similarity.
\newblock {\em IEEE transactions on image processing}, 13(4):600--612, 2004.

\bibitem{wang2003multiscale}
Zhou Wang, Eero~P Simoncelli, and Alan~C Bovik.
\newblock Multiscale structural similarity for image quality assessment.
\newblock In {\em The Thrity-Seventh Asilomar Conference on Signals, Systems \&
  Computers, 2003}, volume~2, pages 1398--1402. Ieee, 2003.

\bibitem{wu2021contrastive}
Haiyan Wu, Yanyun Qu, Shaohui Lin, Jian Zhou, Ruizhi Qiao, Zhizhong Zhang, Yuan
  Xie, and Lizhuang Ma.
\newblock Contrastive learning for compact single image dehazing.
\newblock In {\em Proceedings of the IEEE/CVF Conference on Computer Vision and
  Pattern Recognition}, pages 10551--10560, 2021.

\bibitem{wu2020leed}
Rongliang Wu and Shijian Lu.
\newblock Leed: Label-free expression editing via disentanglement.
\newblock In {\em European Conference on Computer Vision}, pages 781--798.
  Springer, 2020.

\bibitem{wu2020cascade}
Rongliang Wu, Gongjie Zhang, Shijian Lu, and Tao Chen.
\newblock Cascade ef-gan: Progressive facial expression editing with local
  focuses.
\newblock In {\em Proceedings of the IEEE/CVF Conference on Computer Vision and
  Pattern Recognition}, pages 5021--5030, 2020.

\bibitem{wu2018unsupervised}
Zhirong Wu, Yuanjun Xiong, Stella~X Yu, and Dahua Lin.
\newblock Unsupervised feature learning via non-parametric instance
  discrimination.
\newblock In {\em Proceedings of the IEEE conference on computer vision and
  pattern recognition}, pages 3733--3742, 2018.

\bibitem{yu2021dual}
Ning Yu, Guilin Liu, Aysegul Dundar, Andrew Tao, Bryan Catanzaro, Larry~S
  Davis, and Mario Fritz.
\newblock Dual contrastive loss and attention for gans.
\newblock In {\em Proceedings of the IEEE/CVF International Conference on
  Computer Vision}, pages 6731--6742, 2021.

\bibitem{yu2021wavefill}
Yingchen Yu, Fangneng Zhan, Shijian Lu, Jianxiong Pan, Feiying Ma, Xuansong
  Xie, and Chunyan Miao.
\newblock Wavefill: A wavelet-based generation network for image inpainting.
\newblock In {\em Proceedings of the IEEE/CVF International Conference on
  Computer Vision}, pages 14114--14123, 2021.

\bibitem{yu2021diverse}
Yingchen Yu, Fangneng Zhan, Rongliang Wu, Jianxiong Pan, Kaiwen Cui, Shijian
  Lu, Feiying Ma, Xuansong Xie, and Chunyan Miao.
\newblock Diverse image inpainting with bidirectional and autoregressive
  transformers.
\newblock In {\em Proceedings of the 29th ACM International Conference on
  Multimedia}, pages 69--78, 2021.

\bibitem{zhan2021unite}
Fangneng Zhan, Yingchen Yu, Kaiwen Cui, Gongjie Zhang, Shijian Lu, Jianxiong
  Pan, Changgong Zhang, Feiying Ma, Xuansong Xie, and Chunyan Miao.
\newblock Unbalanced feature transport for exemplar-based image translation.
\newblock In {\em Proceedings of the IEEE Conference on Computer Vision and
  Pattern Recognition}, 2021.

\bibitem{zhan2021rabit}
Fangneng Zhan, Yingchen Yu, Rongliang Wu, Kaiwen Cui, Aoran Xiao, Shijian Lu,
  and Ling Shao.
\newblock Bi-level feature alignment for semantic image translation \&
  manipulation.
\newblock {\em arXiv preprint}, 2021.

\bibitem{zhan2021gmlight}
Fangneng Zhan, Yingchen Yu, Rongliang Wu, Changgong Zhang, Shijian Lu, Ling
  Shao, Feiying Ma, and Xuansong Xie.
\newblock Gmlight: Lighting estimation via geometric distribution
  approximation.
\newblock {\em arXiv preprint arXiv:2102.10244}, 2021.

\bibitem{zhan2021multimodal}
Fangneng Zhan, Yingchen Yu, Rongliang Wu, Jiahui Zhang, and Shijian Lu.
\newblock Multimodal image synthesis and editing: A survey.
\newblock {\em arXiv preprint arXiv:2112.13592}, 2021.

\bibitem{zhan2021emlight}
Fangneng Zhan, Changgong Zhang, Yingchen Yu, Yuan Chang, Shijian Lu, Feiying
  Ma, and Xuansong Xie.
\newblock Emlight: Lighting estimation via spherical distribution
  approximation.
\newblock In {\em Proceedings of the AAAI Conference on Artificial
  Intelligence}, pages 3287--3295, 2021.

\bibitem{zhan2019sfgan}
Fangneng Zhan, Hongyuan Zhu, and Shijian Lu.
\newblock Spatial fusion gan for image synthesis.
\newblock In {\em Proceedings of the IEEE conference on computer vision and
  pattern recognition}, pages 3653--3662, 2019.

\bibitem{zhang2021cross}
Han Zhang, Jing~Yu Koh, Jason Baldridge, Honglak Lee, and Yinfei Yang.
\newblock Cross-modal contrastive learning for text-to-image generation.
\newblock In {\em Proceedings of the IEEE/CVF Conference on Computer Vision and
  Pattern Recognition}, pages 833--842, 2021.

\bibitem{zhang2021blind}
Jiahui Zhang, Shijian Lu, Fangneng Zhan, and Yingchen Yu.
\newblock Blind image super-resolution via contrastive representation learning.
\newblock {\em arXiv preprint arXiv:2107.00708}, 2021.

\bibitem{zhang2018lpips}
Richard Zhang, Phillip Isola, Alexei~A Efros, Eli Shechtman, and Oliver Wang.
\newblock The unreasonable effectiveness of deep features as a perceptual
  metric.
\newblock In {\em Proceedings of the IEEE conference on computer vision and
  pattern recognition}, pages 586--595, 2018.

\bibitem{zheng2021spatially}
Chuanxia Zheng, Tat-Jen Cham, and Jianfei Cai.
\newblock The spatially-correlative loss for various image translation tasks.
\newblock In {\em Proceedings of the IEEE/CVF Conference on Computer Vision and
  Pattern Recognition}, pages 16407--16417, 2021.

\bibitem{zhou2017ade20k}
Bolei Zhou, Hang Zhao, Xavier Puig, Sanja Fidler, Adela Barriuso, and Antonio
  Torralba.
\newblock Scene parsing through ade20k dataset.
\newblock In {\em Proceedings of the IEEE conference on computer vision and
  pattern recognition}, pages 633--641, 2017.

\bibitem{zhu2017unpaired}
Jun-Yan Zhu, Taesung Park, Phillip Isola, and Alexei~A Efros.
\newblock Unpaired image-to-image translation using cycle-consistent
  adversarial networks.
\newblock In {\em Proceedings of the IEEE international conference on computer
  vision}, pages 2223--2232, 2017.

\bibitem{zhu2017multimodal}
Jun-Yan Zhu, Richard Zhang, Deepak Pathak, Trevor Darrell, Alexei~A Efros,
  Oliver Wang, and Eli Shechtman.
\newblock Multimodal image-to-image translation by enforcing bi-cycle
  consistency.
\newblock In {\em Advances in neural information processing systems}, pages
  465--476, 2017.

\end{thebibliography}
}

\end{document}